%% file: main.tex
\documentclass{article}

\newif\ifICLR
\ICLRtrue

\usepackage[utf8]{inputenc}
\usepackage{CJKutf8}
\usepackage{tabularx}
\ifICLR
\usepackage{style/iclr2017_workshop}
\else
\usepackage[hyperref]{style/acl2017}
\fi
\usepackage{times}
\usepackage{hyperref}
\usepackage{url}
\usepackage{pgf}
\usepackage{tikz}
\usepackage{amsmath}
\usepackage{amssymb}
\usepackage{amsthm}
\usepackage{outlierpaperdefs}
\usepackage{graphicx}
\usepackage{ctable}
\usepackage{subfig}
\ifICLR
\else
\usepackage{xr-hyper}
\usepackage{hyperref}
\usepackage[capitalise,noabbrev]{cleveref}
\fi
\renewcommand{\tabularxcolumn}[1]{>{\small}m{#1}}
\newcolumntype{Y}{>{\centering\arraybackslash}X}
\input{data/misc_data}

\title{Automated Generation of Multilingual \\ Clusters for the Evaluation of \\ Distributed Representations}
\author{Philip Blair, Yuval Merhav \& Joel Barry \\
Basis Technology\\
One Alewife Center\\
Cambridge, MA 02140 USA \\
\texttt{\{pblair,yuval,joelb\}@basistech.com}
}

\begin{document}

\maketitle

\begin{abstract}
We propose a language-agnostic way of automatically generating sets of semantically similar
clusters of entities along with sets of ``outlier" elements, which may then be used to perform
an intrinsic evaluation of word embeddings in the \textit{outlier detection} task.
We used our methodology to create a 
gold-standard dataset, which we call WikiSem500, and evaluated multiple state-of-the-art embeddings. The results show a
correlation between performance on this dataset and performance on sentiment analysis.
\end{abstract}

\input{sections/introduction}

\input{sections/generating_the_dataset}
\input{sections/evaluation}
\input{sections/future_work}
\input{sections/conclusion}

\bibliography{references}
\ifICLR
\bibliographystyle{style/iclr2017_conference}
\else
\bibliographystyle{style/acl_natbib}
\fi

\ifICLR
\appendix
\input{sections/formalization}
\fi 

\end{document}

%% file: data/misc_data.tex
\newcommand\humanprecision{$68.9\%$}
\newcommand\numhumanresponses{447}

%% file: sections/introduction.tex
\newcommand\ENTITY[2]{{\langle\text{#1},\,\text{\qid{#2}}\rangle}}

\section{Introduction}
High quality datasets for evaluating word and phrase representations are essential for building better models that can advance natural language understanding. Various researchers have developed and shared datasets for syntactic and semantic intrinsic evaluation. The majority of these datasets are based on word similarity (e.g., \citet{finkelstein2001placing,bruni2012distributional,hill2016simlex}) and analogy tasks (e.g., \citet{mikolov2013efficient,mikolov2013linguistic}). While there has been a significant amount of work in this area which has resulted in a large number of publicly available datasets, many researchers have recently identified problems with existing datasets and called for further research on better evaluation methods \citep{faruqui2016problems,gladkova2016intrinsic,hill2016simlex,avraham2016improving,linzen2016issues,batchkarov2016critique}. A significant problem with word similarity tasks is that human bias and subjectivity result in low inter-annotator agreement and, consequently, human performance that is lower than automatic methods \citep{hill2016simlex}. Another issue is low or no correlation between intrinsic and extrinsic evaluation metrics \citep{chiu2016intrinsic,schnabel2015evaluation}.

Recently, \citet{camacho2016find} proposed the \textit{outlier detection task} as an intrinsic evaluation method that improved upon some of the shortcomings of word similarity tasks. The task builds upon the ``word intrusion" task initially described in \citet{Chang:Boyd-Graber:Wang:Gerrish:Blei-2009}: given a set of words, the goal is to identify the word that does not belong in the set. However, like the vast majority of existing datasets, this dataset requires manual annotations that suffer from human subjectivity and bias, and it is not multilingual. 

Inspired by \citet{camacho2016find}, we have created a new outlier detection dataset that can be used for intrinsic evaluation of semantic models. The main advantage of our approach is that it is fully automated using Wikidata and Wikipedia, and it is also diverse in the number of included topics, words and phrases, and languages. At a high-level, our approach is simple: we view Wikidata as a graph, where nodes are entities (e.g., $\ENTITY{Chicago Bulls}{Q128109}$, $\ENTITY{basketball team}{Q13393265}$), edges represent “instance of” and “subclass of” relations (e.g., $\ENTITY{Chicago Bulls}{Q128109}$ is an instance of $\ENTITY{basketball team}{Q13393265}$, $\ENTITY{basketball team}{Q13393265}$ is a subclass of $\ENTITY{sports team}{Q12973014}$), and the semantic similarity between two entities is inversely proportional to their graph distance (e.g., $\ENTITY{Chicago Bulls}{Q128109}$ and  $\ENTITY{Los Angeles Lakers}{Q121783}$ are semantically similar since they are both instance of $\ENTITY{basketball team}{Q13393265}$). This way we can form semantic clusters by picking entities that are members of the same class, and picking outliers with different notions of dissimilarity based on their distance from the cluster entities.   

 
We release the first version of our dataset, which we call WikiSem500, to the research community. It contains around 500 per-language cluster groups for English, Spanish, German, Chinese, and Japanese (a total of 13,314 test cases). While we have not studied yet the correlation between performance on this dataset and various downstream tasks, our results show correlation with sentiment analysis. We hope that this diverse and multilingual dataset will help researchers to advance the state-of-the-art of word and phrase representations.

\section{Related Work}
Word similarity tasks have been popular for evaluating distributional similarity models. 
The basic idea is having annotators assigning similarity scores for word pairs. Models that can automatically assign similarity scores to the same word pairs are evaluated by computing the correlation between their and the human assigned scores. \citet{schnabel2015evaluation} and \citet{hill2016simlex} review many of these datasets. \citet{hill2016simlex} also argue that the predominant gold standards for semantic
evaluation in NLP do not measure the ability of models to reflect similarity. Their main argument is that many such benchmarks measure association and relatedness and not necessarily similarity, which limits their suitability for a wide range of applications. One of their motivating examples is the word pair “coffee” and “cup,” which have high similarity ratings in some benchmarks despite not being very similar. Consequently, they developed guidelines that distinguish between association and similarity and used five hundred Amazon Mechanical Turk annotators to create a new dataset called SimLex-999, which has higher inter annotator agreement than previous datasets. \citet{avraham2016improving} improved this line of work further by redesigning the annotation task from rating scales to ranking, in order to alleviate bias, and also redefined the evaluation measure to penalize models more for making wrong predictions on reliable rankings than on unreliable ones. 

Another popular task is based on word analogies. The analogy dataset
proposed by \citet{mikolov2013efficient} has become a standard
evaluation set. The dataset contains fourteen categories, but
only about half of them are for semantic evaluation (e.g.
``US Cities", ``Common Capitals", ``All Capitals"). In contrast,
WikiSem500 contains hundreds of categories, making it a far more
diverse and challenging dataset for the general-purpose evaluation
of word representations. The Mikolov dataset has the advantage
of additionally including \textit{syntactic} categories, which
we have left for future work.

\citet{camacho2016find} addressed some of the issues mentioned previously by proposing the outlier detection task. Given a set of words, the goal is to identify the word that does not belong in the set. Their pilot dataset consists of eight different topics each made up of a cluster of eight words and eight possible outliers. Four annotators were used for the creation of the dataset. The main advantage of this dataset is its near perfect human performance. However, we believe a major reason for that is the specific choice of clusters and the small size of the dataset. 


%% file: sections/generating_the_dataset.tex
\section{Generating the Dataset}


In a similar format to the one used in the dataset furnished by \citet{camacho2016find}, 
we generated sets of entities which were semantically similar to one another, 
known as a ``cluster", followed by up to three pairs (as available) of dissimilar entities, 
or ``outliers", each with different levels of semantic similarity to the cluster. The core
thesis behind our design is that our knowledge base, \citet{wikidata}, can be treated like 
a graph, where the semantic similarity between two elements is inversely proportional to 
their graph distance. 

\input{figures/cluster_entities}

Informally, we treat Wikidata entities which are instances of a common entity as a cluster (see 
Figure~\ref{fig:wikidata-cluster}). 
Then, starting from that common entity (which we call a `class'), we follow ``subclass of" 
relationships to find a sibling class (see ``American Football Team" in Figure~\ref{fig:wikidata-cluster}). Two items which are instances of the sibling class (but
\textit{not} instances of the original class) are chosen as outliers. The process is then
repeated with a `cousin' class with a common \textit{grandparent} to the original class (see ``Ice Hockey Team" in Figure~\ref{fig:wikidata-cluster}).
Finally, we choose two additional outliers by randomly selecting items which are a distance
of at least 7 steps away from the original class. These three ``outlier classes" are referred
to as $\oone$, $\otwo$, and $\othree$ outlier classes, respectively.

A full formalization of our approach is described in Appendix~\ref{sec:formalization}.

\subsection{Refining the Dataset Quality}
\label{subsec:dataset-quality}
Prior to developing a framework to improve the quality of the generated dataset, we
performed a small amount of manual pruning of our Wikidata graph. Disambiguation pages
led to bizarre clusters of entities, for their associated relationships are not true semantic connections, but are instead artifacts of the structure of our
knowledge base. As such, they were removed. Additionally, classes within a distance of
three from the entity for ``Entity" itself\footnote{\qid{Q35120} is effectively the ``root" node of the Wikidata graph; 95.5\% of nodes have ``subclass of" chains which terminate at this node.} (\qid{Q35120}) had instances which had quite weak semantic similarity
(one example being ``human"). We decided that entities at this depth range ought to
be removed from the Wikidata graph as well.

Once our Wikidata dump was pruned, we employed a few extra steps at generation time to
further improve the quality of the dataset; first and foremost were how we chose representative
instances and outliers for each class (see $\sigma_i$ and $\sigma_o$ in Appendix~\ref{sec:formalization}).
While ``San Antonio Spurs" and ``Chicago Bulls" may both
be instances of ``basketball team", so are ``BC Andorra" and ``Olimpia Milano." We 
wanted the cluster entities to be as strongly related as possible, so we sought a
class-agnostic heuristic to accomplish this. Ultimately, we found that favoring entities
whose associated Wikipedia pages had higher sitelink counts gave us the desired effect.

As such, we created clusters by choosing the top eight instances of a given class, ranked
by sitelink count. Additionally, we only chose items as outliers when they had at least
ten sitelinks 
so as to remove those which were `overly
obscure,' for the ability of word embeddings to identify rare words \citep{schnabel2015evaluation} would artificially decrease the difficulty of such outliers.

We then noticed that many cluster entities had similarities in their labels that could
be removed if a different label was chosen. For example, 80\% of the entities chosen 
for ``association football club" ended with the phrase ``F.C." This 
essentially invalidates the cluster, for the high degree of syntactic overlap artificially increases
the cosine similarity of all cluster items in word-level embeddings. In order to increase
the quality of the surface forms chosen for each entity, we modified our resolution of entity QIDs to surface forms (see $\tau$ in Appendix~\ref{sec:formalization}) to incorporate a variant\footnote{By `variant,' we are referring to the fact that
the dictionaries in which we perform the probability lookups are constructed for
\textit{each language}, as opposed to the cross-lingual dictionaries originally described
by \citet{spitkovsky2012}.} of the work from \citet{spitkovsky2012}:
\begin{equation}
\tau(QID) = \argmax_s\{P(s\mid \operatorname{wikipedia\,page}(QID))\}
\end{equation}
That is, the string for an entity is the string which is most likely to link to the
Wikipedia page associated with that entity. For example, half of the inlinks to the page
for Manchester United FC are the string ``Manchester United," which is the colloquial
way of referring to the team.

Next, we 
filter out remaining clusters using a small set of heuristics. The following clusters are rejected:
\begin{itemize}
    \item Clusters with more than two items are identical after having all digits removed. 
    This handles cases such as entities only differing by years (e.g. ``January 2010," ``January 2012," etc.).
    \item Clusters with more than three elements have identical first or last six characters\footnote{
    For Chinese and Japanese, this is modified such that the at least six entities must
    have identical (non-kana) first or last characters, or more than three must have 
    identical the same first or last \textit{two} characters. Because English is not
    inflected, we simply use spaces as approximate word boundaries and check that the
    first or last of \textit{those} does not occur too often.}. Characters are compared
    instead of words in order to better support inflected languages. This was inspired by
    clusters for classes such as ``counties of Texas" (\qid{Q11774097}), where even the
    dictionary-resolved aliases have high degrees of syntactic overlap (namely, over half
    of the cluster items ended with the word ``County").
    \item Clusters in which any item has an occurrence of a `stop affix,' such as the
    prefix ``Category:" or the suffix ``\jpn{一覧}" (a Japanese Wikipedia equivalent of
    ``List of"). In truth, this could be done during preprocessing, but doing it at
    cluster generation time instead has no bearing on the final results. These were
    originally all included under an additional stop class (``Wikimedia page outside the
    main knowledge tree") at prune time, but miscategorizations in the Wikidata 
    hierarchy prevented us from doing so;
    for example, a now-removed link resulted in every country being pruned from the
    dataset. As such, we opted to take a more conservative approach and perform this
    on at cluster-generation time and fine tune our stoplist as needed.
    \item Clusters with more than one entity with a string length of one. This prevents
    clusters such as ``letters of the alphabet" being created. Note that this heuristic was disabled
    for the creation of Chinese and Japanese clusters.
    \item Clusters with too few entities, after duplicates introduced by resolving entities to surface forms ($\tau$) are removed.
\end{itemize}

\subsection{The WikiSem500 Dataset}
\label{subsec:our-dataset}
Using the above heuristics and preprocessing, we have generated a dataset, which we call
\textbf{WikiSem500}\footnote{The dataset is available for download at \dataseturl}. Our dataset is formatted as a series of files containing \textit{test groups}, comprised of a cluster and a series of outliers. \textit{Test cases} can be constructed by taking each outlier in a given group with that
group's cluster. Table~\ref{tbl:wikisem500-stats} shows the number of included test groups and test cases for each language. Each group contains a cluster of 7-8 entities and up to two entities from each of the three outlier classes. Table~\ref{tbl:sample-clusters} shows example clusters taken from the dataset.

\input{data/wikisem500_stats}

\input{data/sample_clusters}

%% file: figures/cluster_entities.tex
\ifICLR
\newcommand{\figwikidataclusterwidth}{.6\textwidth}
\else
\newcommand{\figwikidataclusterwidth}{\columnwidth}
\fi
\begin{figure}
    \centering
    \includegraphics[width=\figwikidataclusterwidth]{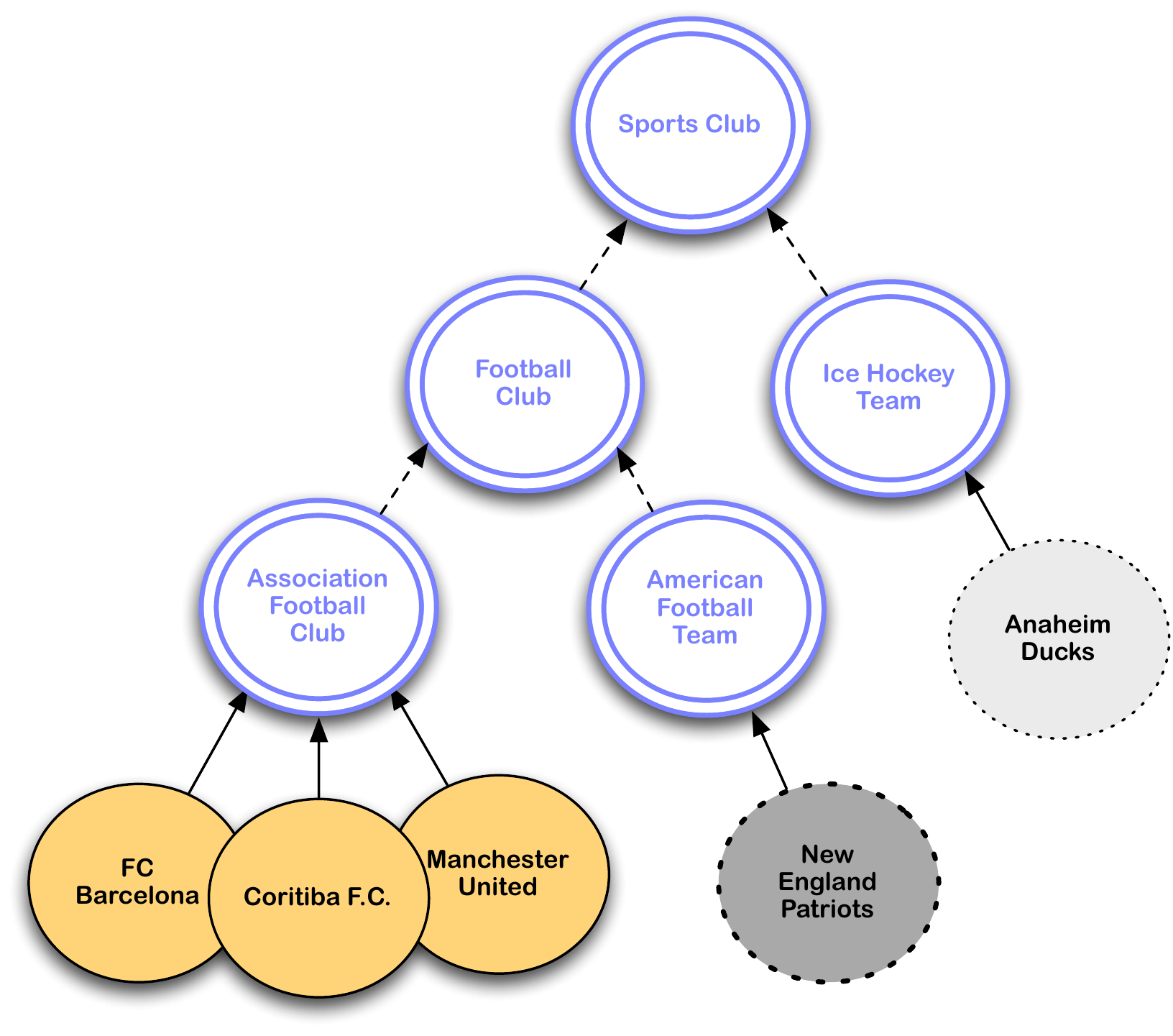}
    \caption{Partial example of a Wikidata cluster. Solid arrows represent ``Instance Of" relationships, and dashed arrows represent ``Subclass Of" relationships.}
    \label{fig:wikidata-cluster}
\end{figure}

%% file: data/wikisem500_stats.tex
\begin{table}
\caption{Statistics of the WikiSem500 dataset.}
\label{tbl:wikisem500-stats}
\begin{center}
\bgroup
\def\arraystretch{1.2}
\begin{tabular}{|c|c|c|}
\hline
\textbf{Language} & \textbf{Test Groups} & \textbf{Test Cases} \\
\hline
English  & 500 & 2,816 \\
Spanish  & 500 & 2,777 \\
German   & 500 & 2,780 \\
Japanese & 448 & 2,492 \\
Chinese  & 441 & 2,449 \\
\hline
\end{tabular}
\egroup
\end{center}
\end{table}

%% file: data/sample_clusters.tex
\begin{table}[t]
\caption{Example partial clusters from the WikiSem500 dataset. Classes, clusters, and outliers are shown.}
\label{tbl:sample-clusters}
\begin{center}
\ifICLR
\bgroup
\def\arraystretch{1.2}
\begin{tabularx}{\textwidth}{|c|*4{Y|}}
     \cline{2-5}
     \multicolumn{1}{c|}{}
     & {\bf fictional country} & {\bf mobile operating system} & {\bf video game publisher} & {\bf emotion} \tabularnewline
     \specialrule{.075em}{.075em}{.075em}
     \multirow{4}{*}{Cluster Items}
                 & Mordor            & Windows Phone   & Activision & fear
     \tabularnewline
     \cline{2-5} & Rohan             & Firefox OS      & Nintendo   & love
     \tabularnewline
     \cline{2-5} & Shire             & iOS       & Valve Corporation & happiness
     \tabularnewline
     \cline{2-5} & Arnor             & Android   & Electronic Arts   & anger
     \tabularnewline
     \specialrule{.075em}{.075em}{.075em}
     \multirow{5}{*}{Outliers}
                 & Thule             & Periscope & HarperCollins     & magnitude
     \tabularnewline
     \cline{2-5} &  Duat             & Ingress   & Random House      & Gini coefficient
     \tabularnewline
     \cline{2-5} & Donkey Kong       &  iWeb  & Death Row Records    & redistribution of wealth
     \tabularnewline
     \cline{2-5} & Scrooge McDuck    & iPhoto & Sun Records & Summa Theologica
     \tabularnewline
     \specialrule{.075em}{.075em}{.075em}
\end{tabularx}
\egroup
\else
\bgroup
\def\arraystretch{1}
\begin{tabularx}{\columnwidth}{|c|*3{Y|}}
     \cline{2-4}
     \multicolumn{1}{c|}{}
     & {\bf fictional country} & {\bf video game publisher} & {\bf emotion} \tabularnewline
     \specialrule{.075em}{.075em}{.075em}
     \multirow{5}{*}{\rot{Cluster Items}}
                 & Mordor            & Activision & fear
     \tabularnewline
     \cline{2-4} & Rohan             & Nintendo   & love
     \tabularnewline
     \cline{2-4} & Shire             & Konami & happiness
     \tabularnewline
     \cline{2-4} & Arnor             & Electronic Arts   & anger
     \tabularnewline
     \specialrule{.075em}{.075em}{.075em}
     \multirow{6}{*}{\rot{Outliers}}
                 & Thule             & HarperCollins     & magnitude
     \tabularnewline
     \cline{2-4} &  Duat             & Random House      & Gini coefficient
     \tabularnewline
     \cline{2-4} & Donkey Kong       & Death Row Records    & redistribution of wealth
     \tabularnewline
     \cline{2-4} & Scrooge McDuck    & Sun Records & Summa Theologica
     \tabularnewline
     \specialrule{.075em}{.075em}{.075em}
\end{tabularx}
\egroup
\fi 
\end{center}
\end{table}

%% file: sections/evaluation.tex
\section{Evaluation}
\label{sec:evaluation}

For clarity, we first restate the definitions of the scoring metrics defined by
\citet{camacho2016find} in terms of test groups (in contrast to the original definition,
which is defined in terms of test \textit{cases}). The way in which out-of-vocabulary
entities are handled and scores are reported makes this distinction important, as will
be seen in Section~\ref{subsec:embed-results}.

The core measure during evaluation is known as the \textit{compactness score}; given a
set $W$ of words, it is defined 
\ifICLR
\else
for a word $w\in W$
\fi
as follows:
\ifICLR
\begin{equation}
\label{eqn:compactness}
\forall w\in W, 
c(w)=
\frac{1}{(\lvert W\rvert-1)(\lvert W\rvert - 2)}
\sum_{w_i\in W\setminus \{w\}}
\sum_{\substack{w_j\in W\setminus\{w\}\\ w_j\neq w_i}}sim(w_i,w_j)
\end{equation}
\else
\begin{equation}
\label{eqn:compactness}
c(w)=
\frac{\displaystyle\sum_{w_i\in W\setminus \{w\}}
\sum_{\substack{w_j\in W\setminus\{w\}\\ w_j\neq w_i}}sim(w_i,w_j)}{(\lvert W\rvert-1)(\lvert W\rvert - 2)}
\end{equation}
\fi
where $sim$ is a vector similarity measure (typically cosine similarity). Note that
\citet{camacho2016find} reduces the asymptotic complexity of $c(w)$ from $O(n^3)$ to
$O(n^2)$. We denote $P(W,w)$ to be the (zero-indexed) position of $w$ in the list of
elements of $W$, sorted by compactness score in descending
order. From this, we can describe the following definition for
\textit{Outlier Position} (OP), where $\langle C,O\rangle$ is a test group and $o\in O$:
\begin{equation}
\label{eqn:op}
OP(C\cup\{o\}) = P(C\cup\{o\}, o)
\end{equation}
This gives rise to the boolean-valued \textit{Outlier Detection} (OD) function:
\begin{equation}
\label{eqn:od}
OD(C\cup\{o\}) = \begin{cases} 
  1 & OP(C\cup\{o\}) = \lvert C\rvert \\
  0 & \textrm{otherwise}
\end{cases}
\end{equation}

Finally, we can now describe the \textit{Outlier Position Percentage} (OPP) and 
\textit{Accuracy} scores:
\begin{equation}
\label{eqn:opp}
OPP(D) = \frac{\sum_{\langle C,O\rangle\in D}\sum_{o\in O}\frac{OP(C\cup\{o\})}{\lvert C\rvert}} {\sum_{\langle C, O\rangle\in D}\lvert O\rvert}
\end{equation}
\begin{equation}
\label{eqn:acc}
Accuracy(D) = \frac{\sum_{\langle C,O\rangle\in D}\sum_{o\in O}OD(C\cup\{o\})} {\sum_{\langle C,O\rangle \in D}\lvert O\rvert}
\end{equation}

\subsection{Handling Out-of-Vocabulary Words}
\label{subsec:oov}
One thing \citet{camacho2016find} does not address is how out-of-vocabulary (OOV)
items should be handled. Because our dataset is much larger and contains a wider variety of words,
we have extended their work to include additional scoring provisions which better encapsulate the performance
of vector sets trained on different corpora.

There are two approaches to handling out-of-vocabulary entities: use a sentinel vector to represent
all such entities or discard such entities entirely. The first approach is simpler, but it has a number
of drawbacks; for one, a poor choice of sentinel can have a drastic impact on results. For example, an
implementation which uses the zero vector as a sentinel and defines 
$sim(\vec{x},\vec{0})=0\,\forall \vec{x}$ places many non-out-of-vocabulary outliers at a large disadvantage
in a number of vector spaces, for we have found that negative compactness scores are rare. The second approach avoids
deliberately introducing invalid data into the testing evaluation, but comparing scores across vector
embeddings with different vocabularies is difficult due to them having different in-vocabulary subsets of the
test set.

We have opted for the latter approach, computing the results on both the entire dataset and on only the 
intersection of in-vocabulary entities between all evaluated vector embeddings. This allows us to compare
embedding performance both when faced with the same unknown data and when evaluated on the same, 
in-vocabulary data.

\subsection{Human Baseline}
\label{subsec:human-baseline}
In order to gauge how well embeddings should perform on our dataset, we conducted a human evaluation.
We asked participants to select the outlier from a given test case, providing us with a human baseline
for the accuracy score on the dataset. We computed the non-out-of-vocabulary intersection of the embeddings
shown in Table~\ref{tbl:intersection}, from which 60 test groups were sampled. Due to the wide array of
domain knowledge needed to perform well on the dataset, participants were allowed to refer to Wikipedia
(but explicitly told \textit{not} to use Wikidata). We collected \numhumanresponses{} responses, with an overall precision 
of \humanprecision. 

The performance found is not as high as on the baseline described in
\citet{camacho2016find}, so we conducted a second human evaluation on a smaller hand-picked set
of clusters in order to determine whether a lack of domain knowledge or a systemic issue with
our method was to blame. We had 6 annotators fully annotate 15 clusters generated with our
system. Each cluster had one outlier, with a third of the clusters having each of the three
outlier classes. Human performance was at 93\%, with each annotator missing exactly one 
cluster. Five out of the six annotators missed the same cluster, which was based on books and
contained an $\oone$ outlier (the most difficult class).
We interviewed the annotators, and three of them cited a lack of clarity on Wikipedia over
whether or not the presented outlier was a book (leading them to guess), while the other two
cited a conflation with one of the book titles and a recently popular Broadway production.

With the exception of this cluster, the performance was near-perfect, with one annotator 
missing one cluster. Consequently, we believe that the lower human performance on our dataset 
is primarily a result of the dataset's broad domain.

\subsection{Embedding Results}
\label{subsec:embed-results}
We evaluated our dataset on a number of publicly available vector embeddings: the Google
News-trained CBOW model released by \citet{mikolov2013efficient}, the 840-billion token Common Crawl corpus-trained GloVe model released by \citet{pennington2014glove}, and the English, Spanish,
German, Japanese, and Chinese MultiCCA vectors\footnote{The vectors are word2vec CBOW vectors,
and the non-English vectors are aligned to the English vector space. Reproducing the original (unaligned) non-English
vectors yields near-identical results to the aligned vectors.} from \citet{ammar2016massively}, which are trained on a combination of the Europarl \citep{koehn2005europarl} and Leipzig \citep{quasthoff2006corpus} corpora. In addition,
we trained GloVe, CBOW, and Skip-Gram \citep{mikolov2013efficient} models on an identical
corpus comprised of an English Wikipedia dump and Gigaword corpus\footnote{We used the July 2016 Wikipedia dump \citep{wikipedia2016dump} and the 2011 Gigaword corpus \citep{parker2011gigaword}.}.

The bulk of the embeddings we evaluated were word embeddings (as opposed to
phrase embeddings), so we needed to combine each embeddings' vectors in order to represent
multi-word entities.
If the embedding \textit{does} handle phrases (only Google News), we perform a greedy lookup for
the longest matching subphrase in the embedding, averaging the subphrase vectors;
otherwise, we take a simple average of the vectors for each token in the phrase. If
a token is out-of-vocabulary, it is ignored.
If all
tokens are out-of-vocabulary, the entity is discarded. This check happens as
a preprocessing step in order to guarantee that a test case does not have its outlier
thrown away. As such, we report the percentage of cluster entities filtered out for
being out-of-vocabulary \textit{separately} from the outliers which are filtered out,
for the latter results in an entire test case being discarded.

In order to compare how well each vector embedding would do when run on unknown input
data, we first collected the scores of each embedding on the entire dataset.
Table~\ref{tbl:english} shows the Outlier Position Percentage (OPP) and accuracy scores 
of each embedding, along with the number of
test groups which were skipped entirely\footnote{This happens when either all outliers
are out-of-vocabulary or fewer than two cluster items are in-vocabulary. 
No meaningful evaluation can be performed on the remaining data, so the group is 
skipped.} and the mean percentage of out-of-vocabulary cluster entities and outliers among 
all test groups\footnote{This includes the out-of-vocabulary rates of the skipped
groups.}. As in \citet{camacho2016find}, we used cosine similarity for the $sim$ measure
in Equation~\ref{eqn:compactness}.

\input{data/english_results}

The MultiCCA (Leipzig+Europarl) CBOW vectors have the highest rate of out-of-vocabulary entities,
likely due in large part to the fact that its vocabulary is an order of magnitude smaller than
the other embeddings (176,691, while the other embeddings had vocabulary sizes of over 1,000,000). Perhaps most surprising is
the below-average performance of the Google News vectors. While attempting to understand
this phenomenon, we noticed that disabling the phrase vectors \textit{boosted} performance; as
such, we have reported the performance of the vectors with and without phrase vectors
enabled. 
%

Inspecting the vocabulary of the Google News vectors, we have inferred that the vocabulary has undergone some form of normalization; performing the normalizations which
we can be reasonably certain were done before evaluating has a negligible impact
($\approx +0.01\%$) on the overall score. The Google News scores shown in
Table~\ref{tbl:english} are \textit{with} the normalization enabled. Ultimately, we hypothesize
that the discrepancy in Google News scores comes down to the training corpus. We observe a
bias in performance on our training set towards Wikipedia-trained vectors (discussed below; see
Table~\ref{tbl:english-corpora-comparison}), and, additionally, we expect that the Google News
corpus did not have the wide regional coverage that Wikidata has, limiting the training exposure
to many of the more niche classes in the training set.

In order to get a better comparison between the embeddings under identical
conditions, we then took the intersection of in-vocabulary entities across \textit{all}
embeddings and reevaluated on this subset. $23.88\%$ of cluster entities and $22.37\%$
of outliers were out-of-vocabulary across all vectors, with $23$ test groups removed
from evaluation. Table~\ref{tbl:intersection} shows the results of this evaluation.

\input{data/intersection_results}

The scores appear to scale roughly linearly when compared to Table~\ref{tbl:english},
but these results serve as a more reliable `apples to apples' comparison of the 
algorithms and training corpora. 

Because Wikidata was the source of the dataset, we analyzed how using Wikipedia as a
training corpus influenced the evaluation results. We trained three GloVe models with
smaller vocabularies: one trained on only Gigaword, one trained on only Wikipedia, and
one trained on both. The results of evaluating on the embeddings' common intersection are
shown in Table~\ref{tbl:english-corpora-comparison}.
We observe a slight ($\approx 3.15\%$ relative change) bias in OPP scores with Wikipedia over Gigaword,
while finding a significantly larger ($\approx 19.12\%$ relative change) bias in accuracy scores.

\input{data/english_corpora_comparison}

We believe that this bias is acceptable, for OPP scores (which we believe to be more
informative) are not as sensitive to the bias and the numerous other factors involved
in embedding generation (model, window size, etc.) can still be compared by controlling
for the training corpora.

Additionally, we wanted to verify that the $\oone$ outlier class (most similar) was the
most difficult to distinguish from the cluster entities, followed by the $\otwo$ and $\othree$ 
classes. We generated three separate datasets, each with only one class of outliers, and
evaluated each embedding on each dataset. Figure~\ref{fig:opp-acc-vs-class} illustrates a strong
positive correlation between outlier class and both OPP scores and accuracy.

\input{figures/performance_by_class}

Finally, we used the non-English MultiCCA vectors \citep{ammar2016massively} to evaluate
the multilingual aspect of our dataset. We expect to see Spanish and German perform
similarly to the English Europarl+Leipzig vectors, for the monolingual training corpora
used to generate them consisted of Spanish and German equivalents of the English training
corpus. Table~\ref{tbl:multilingual} shows the results of the non-English evaluations.

\input{data/multilingual_results}

We observe a high degree of consistency with the results of the English vectors. The
Japanese and Chinese scores are somewhat lower, but this is likely due to their having
smaller training corpora and more limited vocabularies than their counterparts in other
languages.


\subsection{Correlation with Downstream Performance}
\label{subsec:downstream}
In light of recent concerns raised about the correlation between intrinsic word 
embedding evaluations and performance in downstream tasks, we sought to investigate
the correlation between WikiSem500 performance and extrinsic evaluations. We used the
embeddings from \citet{schnabel2015evaluation} and ran the outlier detection task on
them with our dataset.

As a baseline measurement of how well our dataset correlates with
performance on alternative intrinsic tasks, we compared our evaluation 
with the scores reported in \citet{schnabel2015evaluation} on
the well-known analogy task \citep{mikolov2013efficient}. 
Figure~\ref{subfig:analogy} illustrates strong correlations between analogy task performance and our evaluation's OPP scores and accuracy.

\input{figures/extrinsic_correlation}

Figure~\ref{subfig:extrinsic} displays the Pearson's correlation between
the performance of each embedding on the WikiSem500 dataset and the extrinsic scores
of each embedding on noun-phrase chunking and sentiment analysis reported in 
\citet{schnabel2015evaluation}.

Similar to the results seen in the paper, performance on our dataset correlates
strongly with performance on a semantic-based task (sentiment analysis), with
Pearson's correlation coefficients higher than 0.97 for both accuracy and OPP scores.
On the other hand, we observe a weak-to-nonexistent correlation with chunking. This
is expected, however, for the dataset we have constructed consists of items which
differ in semantic meaning; syntactic meaning is not captured by the dataset. It is worth noting the inconsistency between this and the intrinsic results
in Figure~\ref{subfig:analogy}, which indicate a \textit{stronger} correlation with the syntactic subset of the analogy task than its semantic subset. This is expected, for it
agrees with the poor correlation between chunking and intrinsic
performance shown in \citet{schnabel2015evaluation}.

%% file: data/english_results.tex
\ifICLR
\begin{table}[t]
\caption{Performance of English word embeddings on the entire WikiSem500 dataset.}
\label{tbl:english}
\begin{center}
\bgroup
\def\arraystretch{1.5}
\begin{tabularx}{\textwidth}{|l|l|Y|Y|Y|Y|Y|}
\hline
\multicolumn{1}{|c|}{\bf Model} & \multicolumn{1}{c|}{\bf Corpus} & {\bf OPP} & {\bf Acc.} & {\bf Groups Skipped} & {\bf \% Cluster Items OOV} & {\bf \% Outliers OOV}
\\ \specialrule{.1em}{.1em}{.1em}
\multirow{2}{*}{GloVe} & Common Crawl          & 75.53 & 38.57 & 5  &  6.33 &  5.70 \\
                       & Wikipedia+Gigaword    & 79.86 & 50.61 & 2  &  4.25 &  4.02 \\
\hline
\multirow{4}{*}{CBOW}  & Wikipedia+Gigaword    & 84.97 & 55.80 & 2  &  4.25 &  4.02 \\
                       & Google News (phrases) & 63.10 & 22.60 & 6  & 13.68 & 15.02 \\
                       & Google News           & 65.13 & 24.45 & 6  & 13.68 & 15.02 \\
                       & Leipzig+Europarl      & 74.59 & 42.62 & 18 & 22.03 & 19.62 \\
\hline 
           Skip-Gram   & Wikipedia+Gigaword    & 84.44 & 57.66 & 2  &  4.25 &  4.02 \\
\hline
\end{tabularx}
\egroup
\end{center}
\end{table}
\else
\begin{table*}[t]
\caption{Performance of English word embeddings on the entire WikiSem500 dataset.}
\label{tbl:english}
\begin{center}
\bgroup
\def\arraystretch{1.5}
\begin{tabularx}{\textwidth}{|l|l|Y|Y|Y|Y|Y|}
\hline
\multicolumn{1}{|c|}{\bf Model} & \multicolumn{1}{c|}{\bf Corpus} & {\bf OPP} & {\bf Acc.} & {\bf Groups Skipped} & {\bf \% Cluster Items OOV} & {\bf \% Outliers OOV}
\\ \specialrule{.1em}{.1em}{.1em}
\multirow{2}{*}{GloVe} & Common Crawl          & 75.53 & 38.57 & 5  &  6.33 &  5.70 \\
                       & Wikipedia+Gigaword    & 79.86 & 50.61 & 2  &  4.25 &  4.02 \\
\hline
\multirow{4}{*}{CBOW}  & Wikipedia+Gigaword    & 84.97 & 55.80 & 2  &  4.25 &  4.02 \\
                       & Google News (phrases) & 63.10 & 22.60 & 6  & 13.68 & 15.02 \\
                       & Google News           & 65.13 & 24.45 & 6  & 13.68 & 15.02 \\
                       & Leipzig+Europarl      & 74.59 & 42.62 & 18 & 22.03 & 19.62 \\
\hline 
           Skip-Gram   & Wikipedia+Gigaword    & 84.44 & 57.66 & 2  &  4.25 &  4.02 \\
\hline
\end{tabularx}
\egroup
\end{center}
\end{table*}
\fi

%% file: data/intersection_results.tex
\begin{table}[t]
\caption{Performance of English word embeddings on their common in-vocabulary intersection of the WikiSem500 dataset.}
\label{tbl:intersection}
\begin{center}
\ifICLR
\bgroup
\def\arraystretch{1.5}
\begin{tabular}{|l|l|c|c|}
\hline
\multicolumn{1}{|c|}{\bf Model}  &\multicolumn{1}{c|}{\bf Corpus}  &\multicolumn{1}{c|}{\bf OPP}  &\multicolumn{1}{c|}{\bf Acc.}
\\ \specialrule{.1em}{.1em}{.1em}
\multirow{2}{*}{GloVe} &       Common Crawl & 76.73 & 43.25 \\
                       & Wikipedia+Gigaword & 76.19 & 47.69 \\
\hline
\multirow{4}{*}{CBOW} & Wikipedia+Gigaword          & 82.59 & 55.90 \\
                      & Google News (with phrases)  & 63.67 & 24.74 \\
                      & Google News                 & 66.20 & 27.43 \\
                      & Leipzig+Europarl (MultiCCA) & 75.01 & 42.82 \\
\hline Skip-Gram      & Wikipedia+Gigaword          & 82.03 & 56.80 \\
\hline
\end{tabular}
\egroup
\else
\bgroup
\def\arraystretch{1.5}
\begin{tabularx}{\columnwidth}{|c|Y|c|c|}
\hline
\multicolumn{1}{|c|}{\bf Model}  &\multicolumn{1}{c|}{\bf Corpus}  &\multicolumn{1}{c|}{\bf OPP}  &\multicolumn{1}{c|}{\bf Acc.}
\\ \specialrule{.1em}{.1em}{.1em}
\multirow{2}{*}{GloVe} &       Common Crawl & 76.73 & 43.25
                       \tabularnewline
                       & Wikipedia+Gigaword & 76.19 & 47.69
                       \tabularnewline
\hline
\multirow{4}{*}{CBOW} & Wikipedia+Gigaword          & 82.59 & 55.90
                      \tabularnewline
                      & Google News (with phrases)  & 63.67 & 24.74
                      \tabularnewline
                      & Google News                 & 66.20 & 27.43
                      \tabularnewline
                      & Leipzig+Europarl (MultiCCA) & 75.01 & 42.82
                      \tabularnewline
\hline Skip-Gram      & Wikipedia+Gigaword          & 82.03 & 56.80 \\
\hline
\end{tabularx}
\egroup
\fi 
\end{center}
\end{table}

%% file: data/english_corpora_comparison.tex
\begin{table}[t]
\caption{Performance comparison of GloVe vectors trained on different corpora when evaluated on their common in-vocabulary intersection.}
\label{tbl:english-corpora-comparison}
\begin{center}
\bgroup
\def\arraystretch{1.5}
\begin{tabular}{|l|l|c|}
\hline
\multicolumn{1}{|c|}{\bf Corpus}  &\multicolumn{1}{c|}{\bf OPP}  &\multicolumn{1}{c|}{\bf Acc.}
\\ \hline
Wikipedia+Gigaword & 80.03 & 54.43 \\
Wikipedia & 77.39 & 49.95 \\
Gigaword & 76.36 & 45.07 \\
\hline
\end{tabular}
\egroup
\end{center}
\end{table}

%% file: figures/performance_by_class.tex
\begin{figure}
    \centering
    \ifICLR
    \newcommand{\oppaccvsclasswidth}{.5\textwidth}
    \newcommand{\oppaccvsclasswrap}{\hfill}
    \hfill
    \else
    \newcommand{\oppaccvsclasswidth}{.701\columnwidth}
    \newcommand{\oppaccvsclasswrap}{}
    \hspace{1em}
    \fi
    \subfloat[][\label{subfig:opp-class-perf}]{\includegraphics[width=\oppaccvsclasswidth]{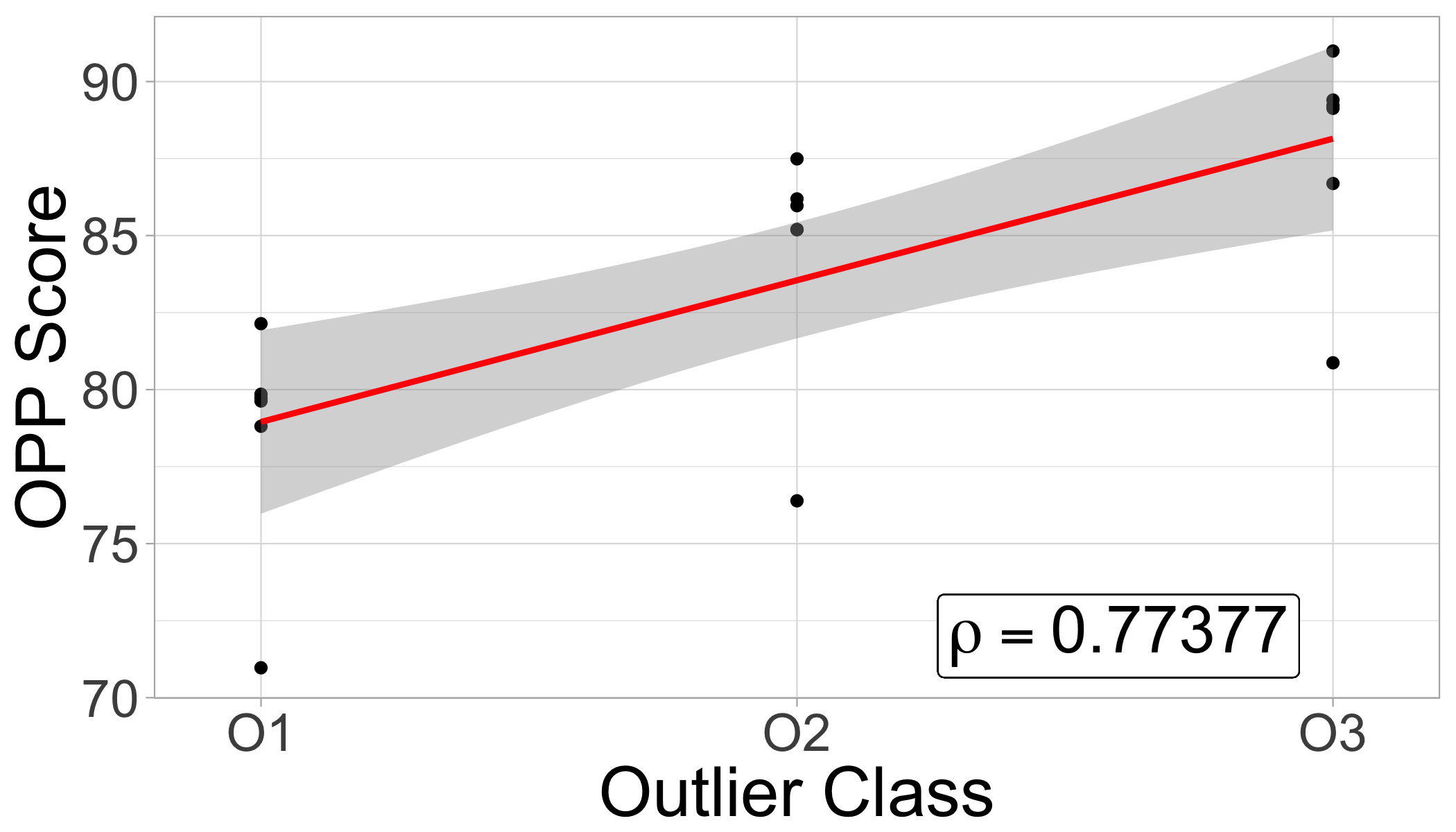}}
    \ifICLR
    \else
    \newline
    \fi
    \subfloat[][\label{subfig:acc-class-perf}]{\includegraphics[width=\oppaccvsclasswidth]{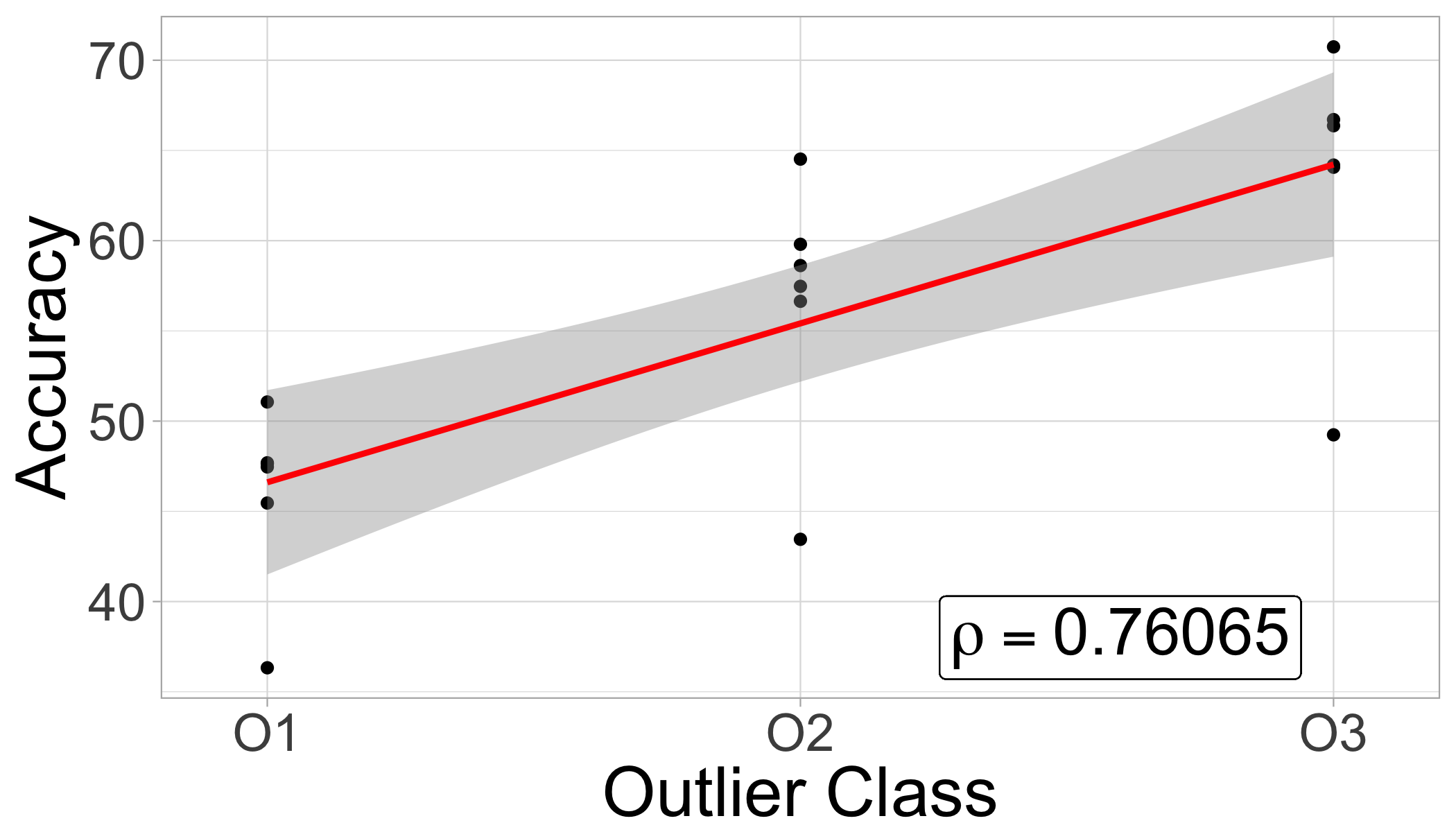}}
    \oppaccvsclasswrap
    \caption{OPP and accuracy scores of embeddings in Table~\ref{tbl:english} by outlier class. The Spearman $\rho$ correlation coefficients are shown.}
    \label{fig:opp-acc-vs-class}
\end{figure}

%% file: data/multilingual_results.tex
\ifICLR
\begin{table}[t]
\caption{Performance of Non-English word embeddings on entire WikiSem500 dataset.}
\label{tbl:multilingual}
\begin{center}
\bgroup
\def\arraystretch{1.5}
\begin{tabularx}{\textwidth}{|l|Y|Y|Y|Y|Y|Y|}
\hline
\multicolumn{1}{|c|}{\bf Language} & {\bf OPP} & {\bf Acc.} & {\bf Groups Skipped} & {\bf \% Cluster Items OOV} & {\bf \% Outliers OOV} & \multicolumn{1}{c|}{\bf Vocab. Size}
\\ \specialrule{.1em}{.1em}{.1em}
       Spanish  & 77.25 & 46.00 & 22 & 21.55 & 17.75 & 225,950 \\
\hline German   & 76.17 & 43.46 & 31 & 24.45 & 25.74 & 376,552 \\
\hline Japanese & 72.51 & 40.18 & 54 & 36.87 & 24.66 & 70,551 \\
\hline Chinese  & 67.61 & 34.58 & 12 & 37.74 & 34.29 & 70,865 \\
\hline
\end{tabularx}
\egroup
\end{center}
\end{table}
\else
\begin{table*}[t]
\caption{Performance of Non-English word embeddings on entire WikiSem500 dataset.}
\label{tbl:multilingual}
\begin{center}
\bgroup
\def\arraystretch{1.5}
\begin{tabularx}{\textwidth}{|l|Y|Y|Y|Y|Y|Y|}
\hline
\multicolumn{1}{|c|}{\bf Language} & {\bf OPP} & {\bf Acc.} & {\bf Groups Skipped} & {\bf \% Cluster Items OOV} & {\bf \% Outliers OOV} & \multicolumn{1}{c|}{\bf Vocab. Size}
\\ \specialrule{.1em}{.1em}{.1em}
       Spanish  & 77.25 & 46.00 & 22 & 21.55 & 17.75 & 225,950 \\
\hline German   & 76.17 & 43.46 & 31 & 24.45 & 25.74 & 376,552 \\
\hline Japanese & 72.51 & 40.18 & 54 & 36.87 & 24.66 & 70,551 \\
\hline Chinese  & 67.61 & 34.58 & 12 & 37.74 & 34.29 & 70,865 \\
\hline
\end{tabularx}
\egroup
\end{center}
\end{table*}
\fi

%% file: figures/extrinsic_correlation.tex
\ifICLR
\begin{figure}
    \centering
    \hfill
    \subfloat[][\label{subfig:analogy}]{\includegraphics[width=.5\textwidth]{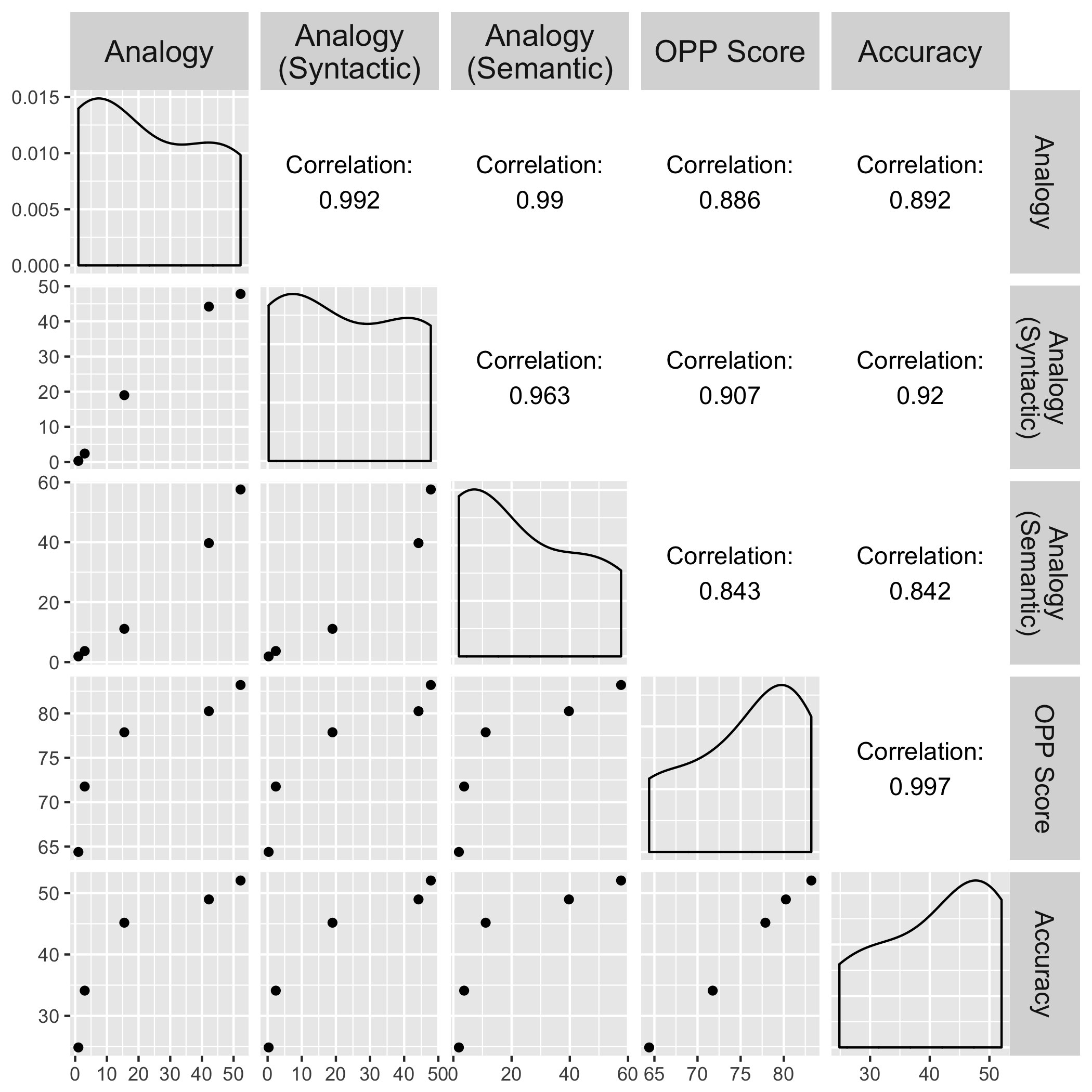}}
    \subfloat[][\label{subfig:extrinsic}]{\includegraphics[width=.5\textwidth]{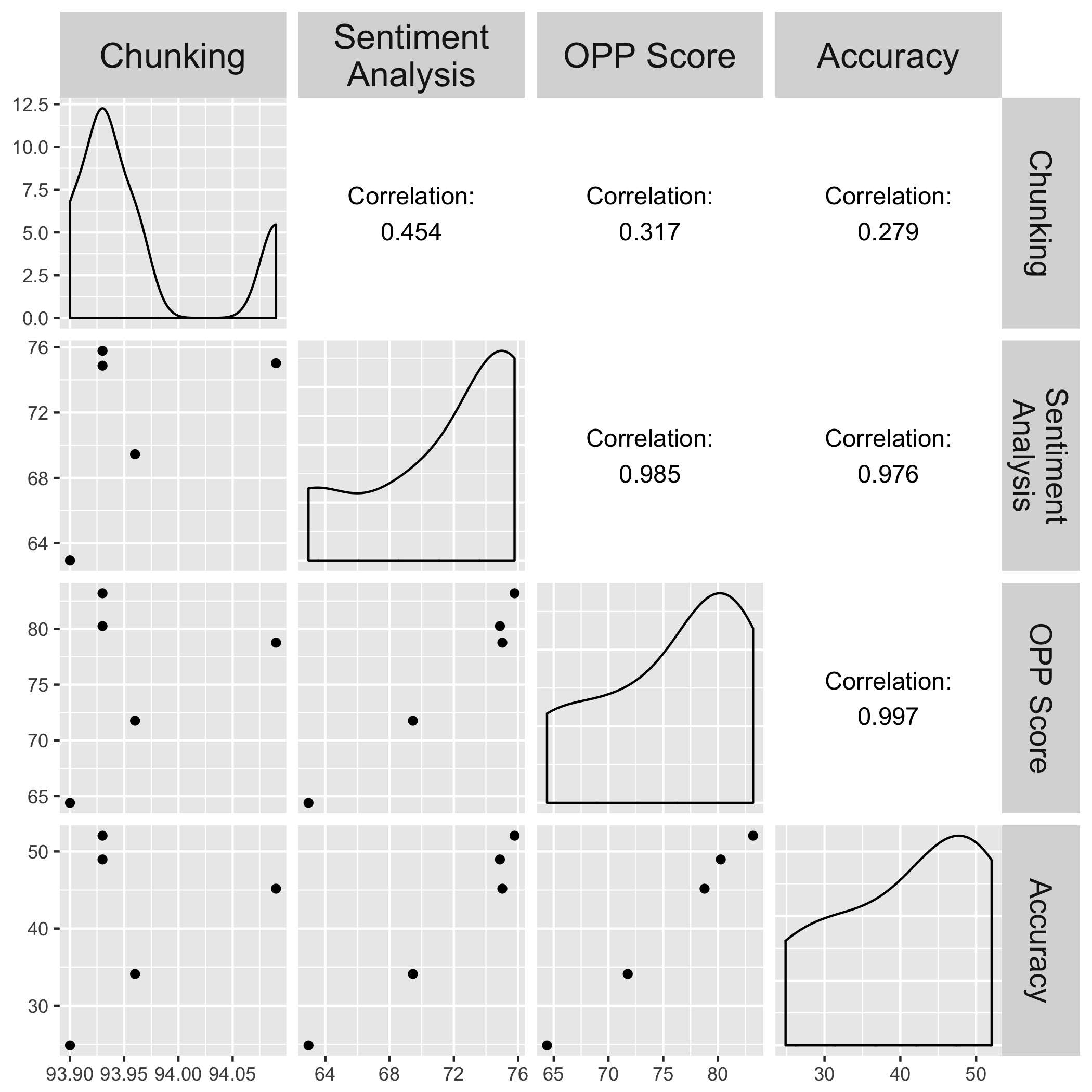}}
    \hfill
    \caption{Pearson's correlation between WikiSem500 outlier detection performance and performance on the analogy task and extrinsic tasks. Distributions of values are shown on the diagonal.}
    \label{fig:extrinsic-correlation}
\end{figure}
\else
\begin{figure*}
    \centering
    \hfill
    \subfloat[][\label{subfig:analogy}]{\includegraphics[width=.5\textwidth]{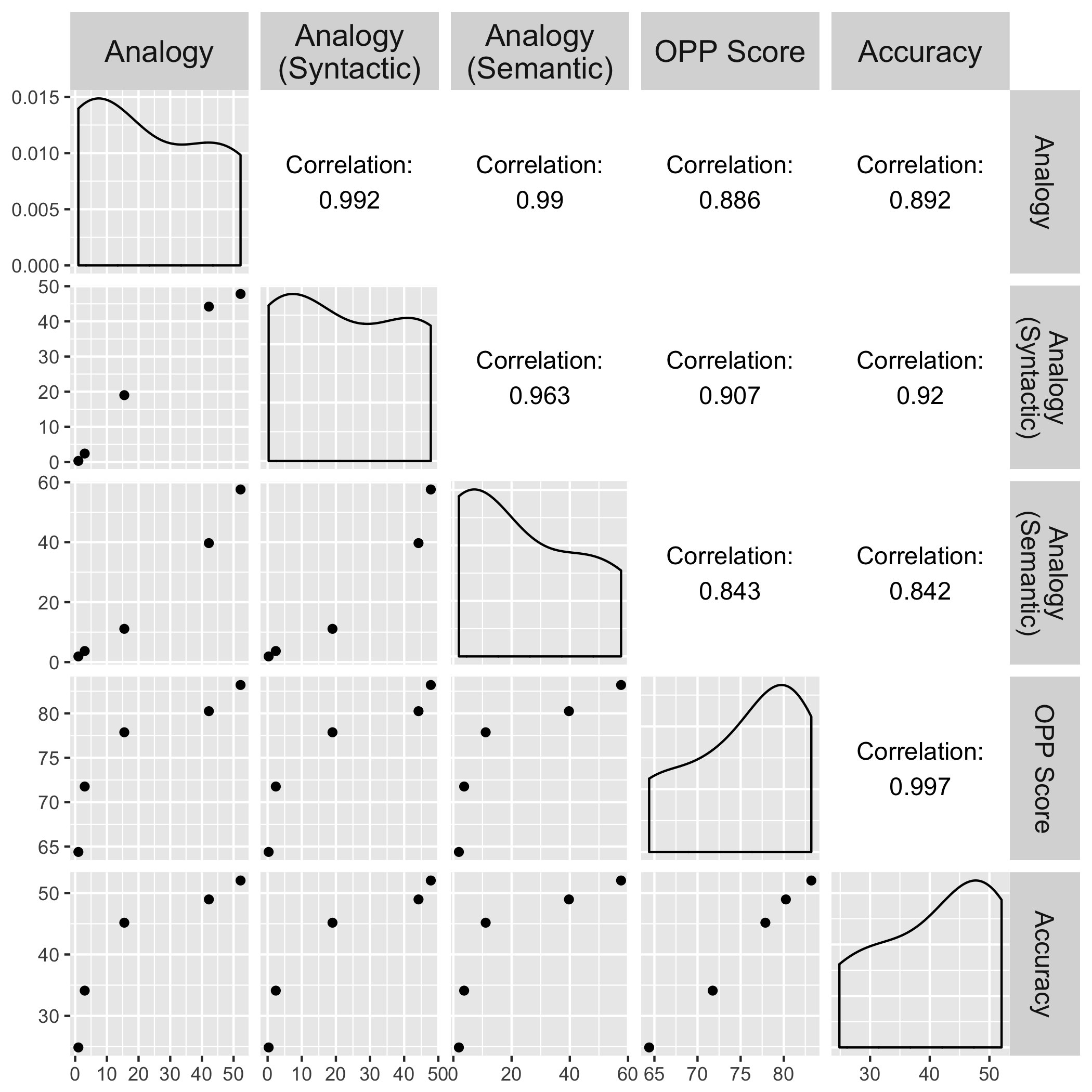}}
    \subfloat[][\label{subfig:extrinsic}]{\includegraphics[width=.5\textwidth]{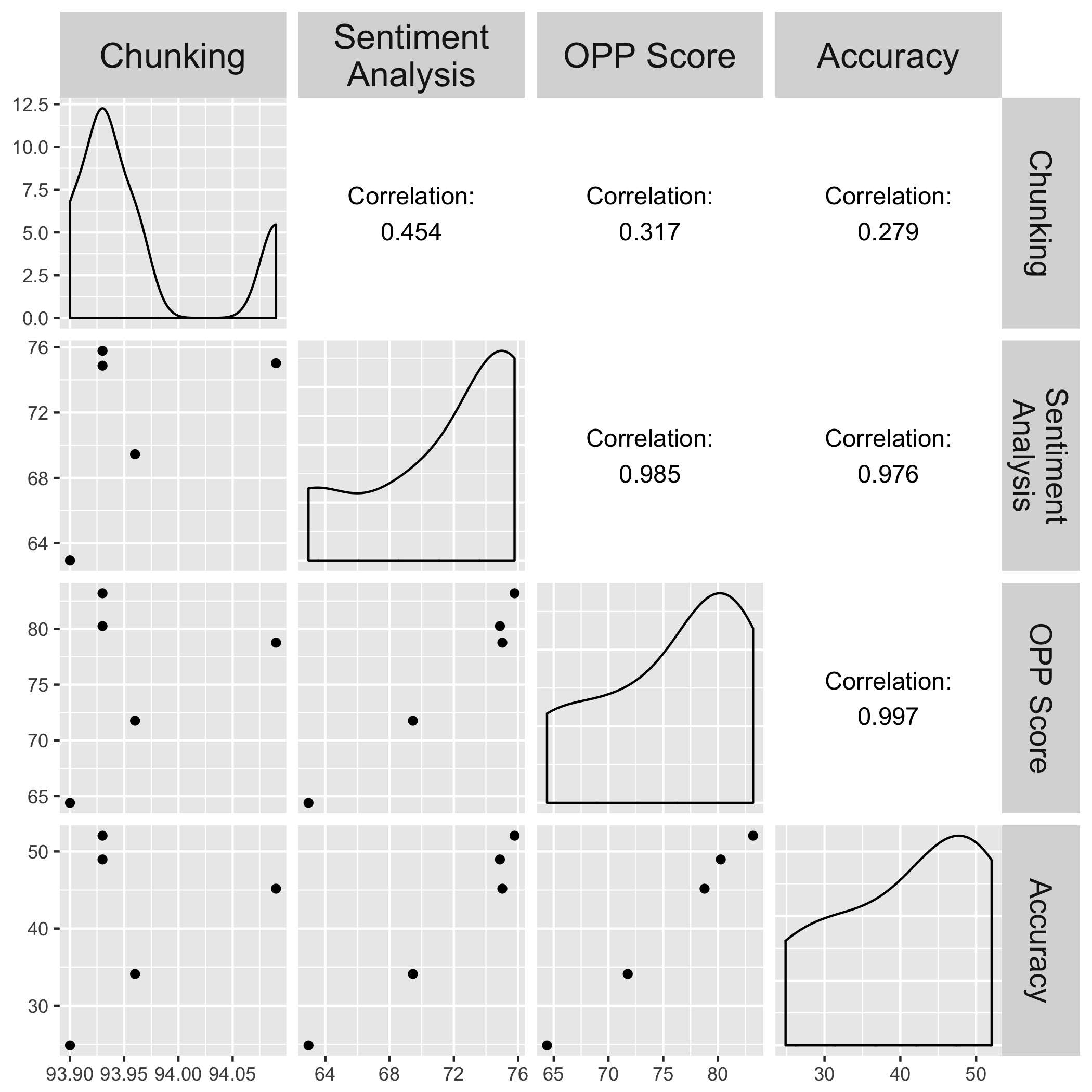}}
    \hfill
    \caption{Pearson's correlation between WikiSem500 outlier detection performance and performance on the analogy task and extrinsic tasks. Distributions of values are shown on the diagonal.}
    \label{fig:extrinsic-correlation}
\end{figure*}
\fi

%% file: sections/future_work.tex
\section{Future Work}
\label{sec:future-work}
Due to the favorable results we have seen from the WikiSem500 dataset, we intend to
release test groups in additional languages using the method described in this paper.
Additionally, we plan to study further the downstream correlation of performance on our dataset
with additional downstream tasks.

Moreover, while we find a substantial correlation between performance on our dataset and on a
semantically-based extrinsic task, the relationship between performance and
syntactically-based tasks leaves much to be desired. We believe that the approach taken
in this paper to construct our dataset could be retrofitted to a system such as
\citet{wordnet} or \citet{wiktionary} (for multilingual data) in order to construct
\textit{syntactically} similar clusters of items in a similar manner. We hypothesize that
performance on such a dataset would correlate much more strongly with
syntactically-based extrinsic evaluations such as chunking and part of speech tagging.

%% file: sections/conclusion.tex
\section{Conclusion}
\label{sec:conclusion}
We have described a language-agnostic technique for generating a dataset consisting of
semantically related items by treating a knowledge base as a graph. In addition, we have
used this approach to construct the \textbf{WikiSem500} dataset, which we have released.
We show that performance on this dataset correlates strongly with downstream performance
on sentiment analysis. This method allows for creation of much larger scale datasets
in a larger variety of languages without the time-intensive task of human creation.
Moreover, the parallel between Wikidata's graph structure and the annotation guidelines
from \citet{camacho2016find} preserve the simple-to-understand structure of the original
dataset.

%% file: sections/formalization.tex
\ifICLR
\else
\documentclass{article}

\usepackage[utf8]{inputenc}
\usepackage{CJKutf8}
\usepackage{tabularx}
\usepackage[hyperref]{style/acl2017}
\usepackage{times}
\usepackage{hyperref}
\usepackage{url}
\usepackage{pgf}
\usepackage{tikz}
\usepackage{amsmath}
\usepackage{amssymb}
\usepackage{amsthm}
\usepackage{outlierpaperdefs}
\usepackage{graphicx}
\usepackage{ctable}
\usepackage{subfig}

\usepackage{xr-hyper}
\usepackage{hyperref}
\usepackage[capitalise,noabbrev]{cleveref}
    \externaldocument[main:]{main}
\renewcommand{\tabularxcolumn}[1]{>{\small}m{#1}}
\newcolumntype{Y}{>{\centering\arraybackslash}X}
\fi

\section{Formalization}
\label{sec:formalization}
We now provide a formal description of the approach taken to generate our dataset.

Let $V$ be the set of entities in Wikidata. For all $v_1, v_2\in V$, we denote the relations
$v_1\instanceof v_2$ when $v_1$ is an instance of $v_2$, and $v_1\subclassof v_2$ when $v_1$ 
is a subclass of $v_2$. We then define $I:V\to V^*$ as the following `instances' mapping:
\begin{equation}
I(v) = \{ v^\prime \in V \mid v^\prime \instanceof v \}
\end{equation}
For convenience, we then denote $C=\{ v \in V \mid \lvert I(v)\rvert \geq 2 \}$; the 
interpretation being that $C$ is the set of entities which have enough instances to
possibly be viable clusters. We now formally state the following definition:
\begin{definition}
A set $A\subseteq V$ is a \textbf{cluster} if $A=I(v)$ for some $v\in C$. We additionally
say that $v$ is the \textbf{class} associated with the cluster $A$.
\label{def:cluster}
\end{definition}
Let $P:V\to V^*$ be the following `parent of' mapping:
\begin{equation}
P(v) = \{ v^\prime \in V \mid v\subclassof v^\prime \}
\end{equation}
Furthermore, let $\inv{P}:V\to V^*$ be the dual of $P$:
\begin{equation}
\inv{P}(v) = \{ v^\prime \in V \mid v^\prime\subclassof v \}
\end{equation}
For additional convenience, we denote the following:
\begin{equation}
P^k(v) = \begin{cases}
P(v) & k = 1\\
\bigcup_{v^\prime\in P(v)} P^{k-1}(v^\prime) & k > 1
\end{cases}
\end{equation}
As an abuse of notation, we define the following:
\begin{equation}
I^*(v) = I(v) \cup \left( \bigcup_{v^\prime \in \inv{P}(v)} I^*(v) \right)
\end{equation}
That is, $I^*(v)$ is the set of all instances of $v$ \textit{and} all instances
of anything that is a subclass of $v$ (recursively). 

We then define the measure $d:V\times V\to\mathbb{N}$ to be the graph distance
between any entities in $V$, using the following set of edges:
\begin{equation}
E_{SU}=\{ (v_1,v_2) \mid v_1 \subclassof v_2 \lor v_2 \subclassof v_1 \}
\end{equation}
Finally, we define\footnote{For the definition of $\otwo$, note that we do
\textit{not} say that it must be true that $p\in P^2(v)\setminus P(v)$. In practice,
however, avoiding (if not excluding) certain values of $p$ in this manner can help
improve the quality of resulting clusters, at the cost of reducing the number of
clusters which can be produced.} three additional
mappings for outliers parametrized\footnote{The WikiSem500 dataset was generated with a value of $\mu=7$.} 
by $\mu\in\mathbb{N}^+$:

\begin{equation}
\oone(v)=\left(\bigcup_{p\in P(v)}\left(\bigcup_{c\in \inv{P}(p)\setminus \{v\}} I^*(c)\right)\right) \setminus I(v)
\end{equation}
\begin{equation}
\otwo(v)=\left(\bigcup_{p\in P^2(v)}\left(\bigcup_{c\in \inv{P}(p)\setminus \{v\}} I^*(c)\right)\right) \setminus I(v)
\end{equation}
\begin{equation}
\othree(v)=\left(\bigcup_{p\in P(v)}\{e\in I(v^\prime)\mid \mu\leq d(p,v^\prime) \}\right) \setminus I(v)
\end{equation}
\ifICLR
\else
\begin{figure*}[t]
\begin{equation}
D = \tau\left(f_D\left(\bigcup_{c\in C} \left\langle f_i(\sigma_i[I(c)]), f_o\left(\sigma_o[\oone(c)] \cup \sigma_o[\otwo(c)] \cup \sigma_o[\othree(c)]\right) \right\rangle\right)\right)
\label{eqn:dataset}
\end{equation}
\end{figure*}
\fi 
To simplify the model, we assume that all three of the above sets are mutually
exclusive. Given these, we can formally state the following definition:
\begin{definition}
Let $A=I(v)$ be a cluster based on a class $v$. An \textbf{outlier} for $A$ is any
$o\in\oone(v)\cup\otwo(v)\cup\othree(v)$. If $o$ is in $\oone(v)$, $\otwo(v)$, or
$\othree(v)$, we denote the \textbf{outlier class} of $o$ as $\oone$, $\otwo$, or 
$\othree$ (respectively).
\label{def:outlier}
\end{definition}
Intuitively, the three outlier classes denote different degrees of `dissimilarity'
from the original cluster; $\oone$ outliers are the most challenging to distinguish,
for they are semantically quite similar to the cluster. $\otwo$ outliers are
slightly easier to distinguish, and $\othree$ outliers should be quite simple to
pick out.

The final dataset (a set of $\langle\textrm{cluster},\textrm{outliers}\rangle$
pairs) is then created by serializing 
\ifICLR
the following:
\begin{equation}
D = \tau\left(f_D\left(\bigcup_{c\in C} \left\langle f_i(\sigma_i[I(c)]), f_o\left(\sigma_o[\oone(c)] \cup \sigma_o[\otwo(c)] \cup \sigma_o[\othree(c)]\right) \right\rangle\right)\right)
\label{eqn:dataset}
\end{equation}
\else
Equation \ref{eqn:dataset}.
\fi 
Where $\sigma_i$ and $\sigma_o$ are functions which select up to a given number
of elements from the given set of instances and outliers (respectively), and
$f_D$, $f_i$, and $f_o$ are functions which filter out dataset elements, instances, 
and outliers (respectively) based on any number of heuristics 
(see Section~\ref{subsec:dataset-quality}). Finally, $\tau$ takes
the resulting tuples and resolves their QIDs to the appropriate surface strings.

The benefit of stating the dataset in the above terms is that it is highly configurable.
In particular, different languages can be targeted by simply changing $\tau$ to resolve
Wikidata entities to their labels in that language. 